\documentclass[10pt,twocolumn,letterpaper]{article}

\usepackage[accsupp]{axessibility} 
\usepackage{multirow}
\usepackage{csquotes}
\usepackage{xcolor}
\usepackage{float}
\usepackage{subcaption}
\usepackage[algorithms]{wacv}

\definecolor{wacvblue}{rgb}{0.21,0.49,0.74}
\usepackage[pagebackref,breaklinks,colorlinks,allcolors=wacvblue]{hyperref}

\usepackage[table]{xcolor}
\usepackage{colortbl}
\newcommand{\best}[1]{\cellcolor[gray]{0.85}\textbf{#1}}
\newcommand{\second}[1]{\cellcolor[gray]{0.93}#1}

\def\wacvPaperID{1285} 
\def\confName{WACV}
\def\confYear{2026}

\title{A Novel Metric for Detecting Memorization in Generative Models\\for Brain MRI Synthesis}

\author{
Antonio Scardace\\
University of Catania\\
{\small{\textbf{joint first authorship}}}\\
{\tt\small antonio.scardace@phd.unict.it}
\and
Lemuel Puglisi\\
University of Catania\\
{\small{\textbf{joint first authorship}}}\\
{\tt\small lemuel.puglisi@phd.unict.it}
\and
Francesco Guarnera\\
University of Catania\\
{\tt\small francesco.guarnera@unict.it}
\and
Sebastiano Battiato\\
University of Catania\\
{\tt\small sebastiano.battiato@unict.it}
\and
Daniele Ravì\\
University of Messina\\
{\tt\small daniele.ravi@unime.it}
}

\begin{document}
\maketitle

\begin{abstract}
Deep generative models have emerged as a transformative tool in medical imaging, offering substantial potential for synthetic data generation. However, recent empirical studies highlight a critical vulnerability: these models can memorize sensitive training data, posing significant risks of unauthorized patient information disclosure. Detecting memorization in generative models remains particularly challenging, necessitating scalable methods capable of identifying training data leakage across large sets of generated samples. In this work, we propose DeepSSIM, a novel self-supervised metric for quantifying memorization in generative models. DeepSSIM is trained to: i) project images into a learned embedding space and ii) force the cosine similarity between embeddings to match the ground-truth Structural Similarity Index (SSIM) scores computed in the image space. To capture domain-specific anatomical features, training incorporates structure-preserving augmentations, allowing DeepSSIM to estimate similarity reliably without requiring precise spatial alignment. We evaluate DeepSSIM in two case studies using synthetic brain MRI and chest X-ray data generated by a Latent Diffusion Model (LDM) trained under memorization-prone conditions. Compared to state-of-the-art memorization metrics, DeepSSIM achieves superior performance, improving F1 scores by an average of +52.03\% over the best existing method. Code and data are publicly available at \url{https://github.com/brAIn-science/DeepSSIM}.
\end{abstract}

\section{Introduction}
Generative models, such as Generative Adversarial Networks (GANs)~\cite{goodfellow2020generative} and diffusion models~\cite{ho2020denoising,rombach2022high}, have become powerful tools in medical imaging, enabling the generation of high-quality synthetic data. A primary objective of these approaches is to produce synthetic scans that preserve the statistical properties of the original data~\cite{pinaya2022brain}. However, concerns have emerged regarding their potential to memorize and inadvertently reproduce private patient data from training datasets, posing serious privacy risks for patients~\cite{somepalli2023diffusion,dar2023investigating}. 

Quantifying memorization in generative models for medical imaging presents both significant technical and conceptual challenges. Existing solutions typically follow a two-step approach: first, they generate a large set of synthetic images and compare them to the training data using a similarity metric; second, they identify memorized training images by checking whether any synthetic image exceeds a predefined similarity threshold. While current methods are designed for generic images and are effective at detecting similarities in broader structural patterns, they struggle to capture the fine-grained anatomical variations crucial for assessing memorization in medical data. For example, brain MRI scans contain highly detailed anatomical structures that can serve as unique identifiers for individual patients~\cite{puglisi2024deepbrainprint}. For a training image to be considered leaked, a generative model must replicate these intricate anatomical details with high fidelity. Assuming that the training and generated images are perfectly aligned, a possible metric for assessing similarity could be the well-known Structural Similarity Index (SSIM)~\cite{wang2004image}. However, SSIM is computationally slow compared to methods that project images into embedding vector spaces. Additionally, the assumption of perfect alignment does not always hold, as this depends on the preprocessing used, limiting the applicability of the SSIM in certain scenarios.

To address this limitation, we introduce DeepSSIM, a novel memorization metric based on SSIM that offers the advantage of providing a scalable, perceptually meaningful similarity measure without requiring image alignment. Additionally, we built a new labeled dataset of similar, different, and duplicate generated scans that can be used to assess memorization metrics. We believe our approach holds potential for assessing privacy risks in generative models trained on medical imaging data, providing a means to safeguard patient confidentiality and offer measurable assurances of privacy preservation.

\section{Related Work} \label{sec:sota}
In this section, we review related work on (i) memorization in generative models, (ii) memorization quantification, and (iii) similarity metrics in medical imaging.

\subsection{Memorization in Generative Models}
The phenomenon of memorization has been examined in generative models, first in GANs~\cite{nagarajan2018theoretical} and more recently in diffusion models~\cite{somepalli2023diffusion}. Early work on membership inference attacks~\cite{chen2020gan} demonstrated that adversaries could determine whether a particular sample was part of the training data. Building on this, extraction attacks pursue a stronger objective aimed at reconstructing sensitive training examples. In particular,~\cite{somepalli2023diffusion} examined memorization in the context of digital forgery, extracting near-exact duplicates of training images from the learned model. Along similar lines,~\cite{carlini2023extracting} conducted a comprehensive analysis demonstrating that state-of-the-art diffusion models are vulnerable to extraction attacks, enabling the recovery of training examples directly from the models. Moreover, their work systematically explored the role of dataset characteristics and training configurations in better understanding the privacy implications of memorization. Another study~\cite{webster2023reproducible} introduced an efficient one-step attack to extract memorized images and proposed the concept of ``template verbatim'', which are training samples that the model reproduces with only minor, non-semantic modifications in specific regions. Complementing these efforts,~\cite{somepalli2023understanding} identified several key factors that exacerbate memorization, including repeated training data, highly specific captions in text-to-image generation, and prolonged training durations. More recently,~\cite{dar2023investigating} investigated memorization in diffusion models trained on medical imaging data, revealing significant privacy risks. Their study showed that models trained on a small dataset of angiography images leaked 59\% of the training data, while models trained on a larger knee MRI dataset still memorized and leaked 33\% of the training set. Collectively, these findings underscore the urgent need to address memorization in generative models, particularly in sensitive domains such as healthcare.

\subsection{Memorization Quantification}
 
Quantifying memorization in generative models is critical for safeguarding patient privacy in medical applications. Prior studies have proposed methods that are tightly coupled with the diffusion model framework, such as those leveraging text-conditional noise prediction~\cite{wen2024detecting} or DDIM inversion~\cite{jiang2025image}. While effective, these approaches rely heavily on specific generative frameworks, which substantially limit their general applicability. An alternative perspective conceptualizes memorization as a form of duplication, thereby linking this problem to well-studied domains such as data contamination~\cite{shi2024detecting} and duplicate detection~\cite{abbas2023semdedup,pizzi2022self}. For instance, SemDeDup~\cite{abbas2023semdedup} identifies duplicate data by embedding items with a foundational model (e.g., CLIP~\cite{radford2021learning}), clustering them in the embedding space, and classifying pairs in the same cluster as duplicates when their similarity exceeds a predefined threshold. Although originally developed for duplicate detection, this approach can be readily adapted to quantify the extent of memorization in generative models.
In~\cite{somepalli2023understanding}, memorization is assessed by embedding both real and synthetic images using the pretrained SSCD (Self-Supervised for image Copy Detection) model~\cite{pizzi2022self}, and computing cosine similarities between the resulting embeddings. The 95th percentile of the similarity distribution is used as the memorization score. In~\cite{chen2024towards}, the authors note that this strategy may underestimate memorization in heavy-tailed distributions. Consequently, they suggest using the maximum similarity scores to capture worst-case memorization scenarios. Similarly,~\cite{chen2024extracting} introduced two new metrics: the Average Memorization Score (AMS), which stratifies image pairs by different levels of similarity, and the Unique Memorization Score (UMS), which measures distinct memorized instances by identifying unique matches—thus accounting for the diversity of memorized images. The authors classify pairs of images into one of the similarity classes by defining arbitrary thresholds. All these methods~\cite{somepalli2023understanding,chen2024extracting,chen2024towards} are built on SSCD, which is pretrained on generic images and not designed to capture the fine-grained details of medical data. As an alternative,~\cite{dar2023investigating} proposes training the feature extractor using contrastive learning directly on the same training set as the generative model under study. This avoids reliance on a pretrained model from a different domain, which could impair performance. \newline

\subsection{Similarity Metrics in Medical Imaging}

To better contextualize the motivation behind DeepSSIM, it is important to consider recent evaluations in the field of Image Quality Assessment (IQA) for medical imaging. Breger et al.~\cite{iqa_measures} conducted a comprehensive analysis of commonly used IQA metrics, including SSIM~\cite{wang2004image} and its variants. Notably, one of their experiments examined SSIM's behavior by comparing original brain MRI scans with lower-quality reconstructions, where degradation was simulated by progressively masking the k-space during reconstruction. Their results revealed that SSIM scores remained relatively stable even as image quality visibly deteriorated. While this insensitivity to degradation is a limitation for IQA tasks, it may be advantageous in applications such as duplicate detection, where robustness to generative artifacts is beneficial. However, duplicate detection also requires millions of comparisons between real and synthetic images, demanding scalability from the similarity metric. SSIM and its variants are poorly suited for such large-scale comparisons due to their computational inefficiency. This underscores the need for metrics like DeepSSIM, which combine the ability to capture fine-grained structural similarities with high computational efficiency and scalability, enabling practical use in high-throughput settings.

\begin{figure*}[t]
\centering
\includegraphics[width=\textwidth]{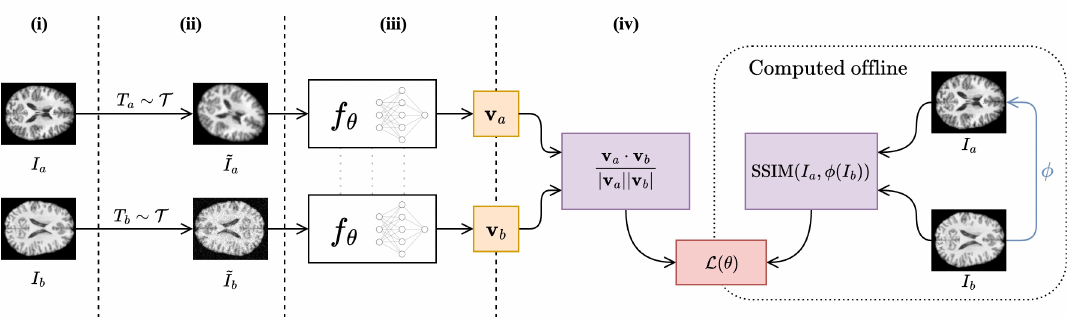}
\caption{
The figure illustrates the DeepSSIM self-supervised training process, which is divided as follows: \textbf{(i)} Inputs: real and synthetic MRI scans are first preprocessed.
\textbf{(ii)} Augmentation: random augmentation to the images is used before feature extraction to make the process robust to image variation.
\textbf{(iii)} Embeddings: images are mapped into a lower-dimensional embedding space using a feature extractor $ f_\theta $.
\textbf{(iv)} Model Optimization: the cosine similarity between the embeddings is directly compared with the ground-truth SSIM to compute the loss and optimize the parameter $\theta $ of our model.
}

\label{fig:pipeline}
\end{figure*}

\section{Methods}
\label{sec:methods}
In this section, we present the two principal components of our DeepSSIM pipeline for detecting memorization. The first component, shown in Figure~\ref{fig:pipeline}, is a self-supervised model that embeds images into a lower-dimensional space (block \textbf{iii} in the figure). This model is explicitly trained to ensure that the cosine similarity between embeddings reflects the original structural similarity of the images, as measured by the ground-truth SSIM scores (\textbf{iv} block in the figure). Next, we define a thresholding technique to determine whether two images are classified as \textit{duplicates}, \textit{similar}, or \textit{different}, based on the similarity of their corresponding embeddings. Finally, we describe the construction and labeling of a dataset comprising paired real and synthetic images, which are used to evaluate the proposed method.

\subsection{Proposed Self-Supervised Similarity Metric}
Let $ \mathcal{P} = \{I_j\} $ denote our training set, we randomly sample a subset $ Q \subset \mathcal{P} \times \mathcal{P} $ of paired images. For each pair $ (I_a, I_b) $, where $ I_a $ is a real image and $ I_b $ is a synthetic image, we perform the following steps: (i) we register $ I_b $ to $ I_a $ using a rigid transformation $\phi$, and (ii) we compute the SSIM score (normalized by brightness) between the aligned images:

\begin{equation}
s_{ab} = \text{SSIM}(I_a, \phi(I_b)).
\end{equation} 

Next, we train a neural network $f_\theta$ to project the images into an embedding space such that their cosine similarity matches their SSIM $ s_{ab} $ in the image space. To improve generalization, we apply random augmentations before extracting features. Specifically, we generate two augmented views of each image:

\begin{equation}
\tilde{I}_a = T_a(I_a), \quad
 \tilde{I}_b = T_b(I_b), \quad
 T_a, T_b \sim \mathcal{T}, 
\end{equation}

where $ \mathcal{T} $ is a set of four random augmentations that preserve anatomical structures. Transformations lacking this property may map distinct anatomies to similar embeddings, resulting in misleading similarity estimates. Specifically, we apply random vertical flips, random horizontal flips, random rotations, and random contrast shifting. The network then maps both images into the embedding space:

\begin{equation}
 \mathbf{v}_a = f_\theta(\tilde{I}_a), \quad \mathbf{v}_b = f_\theta(\tilde{I}_b).
\end{equation}

We optimize $ f_\theta $ by minimizing the mean squared error between the cosine similarity of the embeddings and the ground-truth SSIM score:

\begin{equation}
 \mathcal{L}(\theta) =
 \left\lVert
 \frac{\mathbf{v}_a \cdot \mathbf{v}_b}{|\mathbf{v}_a| |\mathbf{v}_b|} - s_{ab}
 \right\rVert_2.
\end{equation}

Finally, the loss is averaged over a batch of $B$ image pairs. We summarize the entire process of embedding the images and computing cosine similarity as $ s_\theta(I_a, I_b) $.

\subsection{Proposed Memorization Metric}
Let $\mathcal{D}_{train}$ be the training set used to train the diffusion model, and let $S = \{I_b\}$ be a large set of synthetic images generated by this model, with $|S| \gg |\mathcal{D}_{train}|$ to ensure sufficient data for detecting memorization instances. 
\newline

We develop our memorization metric with the aim of quantifying the percentage of training images memorized by the model. To achieve this, we use the previously defined similarity metric $s_\theta$ to compute the similarities between each training image $I_a \in \mathcal{D}_{train}$ and each synthetic image $I_b \in S$. We then classify the pair $(I_a, I_b)$ using the following thresholding function $\psi$:

\begin{equation}
\psi(I_a, I_b) = \begin{cases}
\text{different} & \text{if } s_\theta(I_a,I_b) < \alpha \\
\text{similar} & \text{if } \alpha \leq s_\theta(I_a,I_b) < \beta \\
\text{duplicate} & \text{if } s_\theta(I_a,I_b) \geq \beta
\end{cases}
\end{equation}

where $\alpha$ and $\beta$ are hyperparameters of the pipeline. We define \textit{similar} images as those that display high structural resemblance with minor anatomical variations, whereas \textit{duplicate} images are nearly indistinguishable, differing only in non-semantic factors such as noise, intensity inhomogeneities, or affine transformations. Images that do not meet either criterion are classified as \textit{different}. Finally, we define the memorization metric as the percentage of training images for which at least one duplicate is found in $S$.

\subsection{Memorization-Induced LDM}\label{sec:ldm}
To investigate memorization in generative models, we use a widely adopted Latent Diffusion Model (LDM)~\cite{rombach2022high} as a case study. We train this LDM using a set of real MRI scans (see section~\ref{sec:data}). The LDM is trained to generate 2D axial slices of brain MRIs. To increase exposure to memorization, we condition the image generation process on text prompts specifically tailored to each image, as demonstrated in~\cite{somepalli2023understanding}. These prompts follow the structure below:

\begin{displayquote}
``\textit{middle axial slice from a \texttt{<MRI sequence>} brain MRI (image ID \texttt{<MRI ID>}) acquired from a \texttt{<age>}-year-old \texttt{<biological sex>}, with subject ID \texttt{<subject ID>}, from the \texttt{<dataset name>} dataset.}''
\end{displayquote}

We then use PubMedBERT~\cite{gu2021domain} to embed the text prompt described above and incorporate this embedding into the LDM via a cross-attention layer. Prior findings~\cite{somepalli2023diffusion,carlini2023extracting} also suggest that data duplication in the training set is a key factor in triggering memorization. Therefore, we randomly create between 1 and 15 copies of each training sample.

\subsection{Dataset} \label{sec:data}
We construct two datasets to train and evaluate our system. The first dataset consists of real brain MRI scans and is used to train the LDM. The second consists of real–synthetic image pairs, specifically designed for training and evaluating our memorization detection model, \textbf{DeepSSIM}. Synthetic images are generated by the trained LDM, while real images are taken from the original brain MRI dataset.

\subsubsection{Real MRI Dataset}
The real dataset includes 2,024 T1-weighted and 559 T2-weighted brain MRI scans, totaling 2,583 examples. These scans are sourced from two publicly available datasets: IXI (1,122 scans) and CoRR (1,461 scans). Each 3D scan is processed using a standardized pipeline consisting of N4 bias-field correction, skull stripping, affine registration to MNI space, and intensity normalization. From each volume, we extract the central axial 2D slice, forming the final image dataset $\mathcal{D} = \{ I_i \}_{i=1}^{N}$, where $N = 2,583$. The dataset is used to train the LDM and is partitioned into 85\% training and 15\% validation sets, with subject-level separation to avoid data leakage. The validation set is used for hyperparameter tuning and early stopping during LDM training.

\subsubsection{Real-Synthetic Pair Dataset}\label{sec:synthset}
The second dataset consists of 144,540,750 pairs, each comprising a real brain MRI slice from $\mathcal{D}$ and a corresponding synthetic slice generated by our trained LDM. Specifically, for every real image $I_r \in \mathcal{D}$, we generate 30 synthetic images conditioned on the associated text prompt of the LDM (see Section~\ref{sec:ldm}), yielding a synthetic dataset $S$ with $|S| = |\mathcal{D}| \times 30$. We define the full set of image pairs as $\mathcal{P} = \{ (I_r, I_s) \mid I_r \in \mathcal{D},\ I_s \in S \}$, covering three similarity categories: \textit{different}, \textit{similar}, and \textit{duplicate}. To train the DeepSSIM feature extractor, we randomly sample pairs from this dataset. To evaluate our approach, we construct a dedicated test set designed to capture a balanced and representative distribution of image similarities. Pairing all real images with their synthetic counterparts would result in a highly imbalanced dataset, containing very few duplicate cases. To mitigate this, we curate a balanced test set using a hybrid strategy. In particular, for each real image, 6 synthetic counterparts are selected: i) 3 with the highest FSIM~\cite{fsim_5705575} scores (to increase the chance of selecting duplicates and challenging cases); ii) 3 sampled uniformly at random to ensure diversity. This strategy provides a fair evaluation across a spectrum of difficulty levels and ensures the test set is both challenging and representative. We choose FSIM~\cite{fsim_5705575} as an independent similarity metric for candidate selection to avoid potential bias in the evaluation. FSIM is, in fact, based on fundamentally different principles compared to the ground-truth SSIM and the other baselines, ensuring that candidate selection remains unbiased and does not unfairly favor DeepSSIM or any competing approach.Three expert annotators reviewed the curated test set, providing ground-truth similarity labels for a total of 13,170 pairs. The training set does not require manual labeling, as DeepSSIM is trained via self-supervised learning, using SSIM as a proxy for anatomical similarity. 

\subsection{Pipeline Settings}\label{pipelinesettings}
The embedding neural network $f_\theta$ was instantiated as a ConvNext Base model, with the classification layer replaced by the identity function, and with the first layers frozen. The embedding dimension was set to $256$. The model was trained for $40$ epochs, using the AdamW optimizer with a learning rate of $1 \times 10^{-3}$, a weight decay of $1 \times 10^{-3}$, and a batch size $B$=32. Threshold hyperparameters were set to $\alpha = 0.6$ and $\beta = 0.85$, selected via grid search on the validation set within the range $[0, 1]$ and with a step size of 0.05. The LDM was trained following the protocol proposed in~\cite{pinaya2022brain}, adapted to 2D slices of brain MRIs.

\begin{table*}[t]
\centering
\setlength{\tabcolsep}{5pt}
\def\arraystretch{1.25}

\resizebox{0.95\textwidth}{!} {
\begin{tabular}{|c|cc|cc|cc|c|c|}
\hline
\multirow{2}{*}{\textbf{Method}} 
& \multicolumn{2}{c|}{\textbf{Different}} 
& \multicolumn{2}{c|}{\textbf{Similar}} 
& \multicolumn{2}{c|}{\textbf{Duplicate}} 
& \multirow{2}{*}{\textbf{Macro F1 score}} 
& \multirow{2}{*}{\textbf{Silhouette Score}} \\
\cline{2-7}
& \textbf{Precision} & \textbf{Recall} 
& \textbf{Precision} & \textbf{Recall} 
& \textbf{Precision} & \textbf{Recall} 
& & \\
\hline

\multicolumn{9}{|c|}{\textbf{(A) Performance on Aligned Pairs}} \\
\hline
Chen et al.~\cite{chen2024extracting} & \best{100.00\%} & 27.76\% & 0.00\% & 0.00\% & 29.13\% & \best{100.00\%} & 29.52\% & 0.06 \\
Dar et al.~\cite{dar2023investigating} & 59.32\% & \second{99.97\%} & N/A & N/A & 82.89\% & 1.78\% & 25.98\% & 0.09 \\
SemDeDup~\cite{abbas2023semdedup} & 59.58\% & 99.95\% & 14.48\% & 0.77\% & 0.00\% & 0.00\% & 25.37\% & -0.10 \\
SSIM & \second{99.98\%} & \best{100.00\%} & \best{99.96\%} & \best{99.96\%} & \best{100.00\%} & \second{99.97\%} & \best{99.98\%} & \best{0.54} \\
DeepSSIM & 97.44\% & 82.61\% & \second{54.74\%} & \second{79.93\%} & \second{88.73\%} & 91.82\% & \second{81.55\%} & \second{0.35} \\
\hline

\multicolumn{9}{|c|}{\textbf{(B) Performance on Spatially Misaligned Pairs}} \\
\hline
Chen et al.~\cite{chen2024extracting} & \best{100.00\%} & 27.93\% & 0.83\% & 0.48\% & 31.63\% & \best{99.91\%} & \second{30.78\%} & -0.04 \\
Dar et al.~\cite{dar2023investigating} & 59.21\% & \best{100.00\%} & N/A & N/A & 80.43\% & 1.04\% & 25.48\% & \second{0.05} \\
SemDeDup~\cite{abbas2023semdedup} & 59.81\% & 99.36\% & 22.26\% & 2.40\% & 0.00\% & 0.00\% & 26.33\% & -0.12 \\
SSIM & 61.44\% & \second{99.97\%} & \second{35.48\%} & \second{7.10\%} & \best{94.73\%} & 1.53\% & 30.32\% & 0.04 \\
DeepSSIM & \second{98.15\%} & 81.86\% & \best{53.82\%} & \best{80.41\%} & \second{87.36\%} & \second{91.20\%} & \best{81.00\%} & \best{0.34} \\
\hline

\end{tabular}
}
\caption{\label{tab:brainmri} 
 Quantitative performance evaluation using precision, recall, macro F1 score, and Silhouette Score on \textbf{Brain MRI} data. Results are reported for: \textbf{(A) Aligned Pairs}, consisting of perfectly aligned image pairs; and \textbf{(B) Spatially Misaligned Pairs}, consisting of augmented images with simulated real-world spatial variability.}
\end{table*}

\section{Experiments}

In this section, we conduct five experiments: (i) an ablation study to assess the impact of different backbone architectures on feature extraction performance, (ii) benchmarking our memorization metric against state-of-the-art baseline methods, (iii) evaluation of the generalization capability of our method on cross-modality chest X-ray data, (iv) assessment of the sensitivity of the selected classification thresholds, and (v) measurement and comparison of the computational runtime of our approach against SSIM.

\subsection{Ablation Study}
We evaluate several pre-trained backbone architectures as feature extractors, all trained under the same protocols and tested on a validation set. Our primary evaluation metric is the Mean Absolute Error (MAE) between the cosine similarity of the embeddings and the ground-truth SSIM. The results are reported in Table~\ref{tab:feature_extractors}. We compare four ResNet variants (ResNet-18/34/50/101) alongside the ConvNeXt Base~\cite{Liu_2022_CVPR} architecture. Increasing the ResNet depth does not produce a consistent performance trend, whereas ConvNeXt achieves the lowest MAE, reducing the error by 43\% relative to the best ResNet. These results suggest that ConvNeXt more effectively captures rich, multi-scale features required for perceptual similarity estimation in medical imaging. From this point onward, we adopt ConvNeXt Base as the default backbone for all experiments.

\begin{table}[H]
\centering
\begin{tabular}{lc}
\toprule
\textbf{Pre-trained Backbone} & \textbf{MAE} \\
\midrule
ResNet18& 0.09 $\pm$ 0.008 \\
ResNet34& 0.07 $\pm$ 0.009 \\
ResNet50& \second{0.07 $\pm$ 0.007} \\
ResNet101 & 0.09 $\pm$ 0.009 \\
ConvNeXt Base~\cite{Liu_2022_CVPR} & \best{0.04 $\pm$ 0.009} \\
\bottomrule
\end{tabular}
\caption{\label{tab:feature_extractors}Comparison of pre-trained backbone architectures within the DeepSSIM pipeline. The MAE is calculated between the predicted similarity and the ground-truth SSIM.}
\end{table}

\subsection{Quantitative Comparison}
We perform a quantitative evaluation by classifying each image pair into the corresponding classes and reporting precision, recall, and macro F1 score for each method. Additionally, we calculate the Silhouette Score~\cite{rousseeuw1987silhouettes} to assess how well the classes are separated, offering a performance measure independent of threshold choices. This experiment is conducted under two distinct settings: (A) an ideal setting with perfectly aligned image pairs, and (B) a more challenging setting where image pairs undergo spatial variations. In the latter scenario, images in all pairs are subject to random anatomy-preserving transformations such as horizontal flip, vertical flip, random rotation, and contrast adjustment.

\subsubsection{Baselines}

We assess the performance of DeepSSIM against state-of-the-art methods using the labeled dataset described in Section~\ref{sec:synthset}. The first baseline metric is UMS, introduced by Chen et al.~\cite{chen2024extracting}. This approach classifies pairs into four categories: different, low, medium, or high similarity. To adapt it to our setting, we combined the low and medium similarity categories into a single \textit{similar} class and treated high similarity as \textit{duplicate}. The second baseline is the contrastive method of Dar et al.~\cite{dar2023investigating}, which uses a single cutoff at the 95th percentile of training–validation similarity scores. As a result, it distinguishes only duplicates from non-duplicates and does not explicitly identify similar pairs. As the official code for this approach was not available, we replicated the method based on the details provided in the paper, following the original setup as closely as possible. The final baseline in our evaluation is SemDeDup~\cite{abbas2023semdedup}. We aligned its classification scheme with ours by mapping the ``perceptual duplicate'' category to our \textit{duplicate} class and the ``semantic duplicate'' category to our \textit{similar} class. To better adapt the method to the medical domain, we use BiomedCLIP~\cite{zhang2023biomedclip} as the underlying foundational model in SemDeDup. For completeness, we also compare DeepSSIM against a standard SSIM baseline. 

\begin{figure}[t]
\centering
\includegraphics[width=\linewidth]{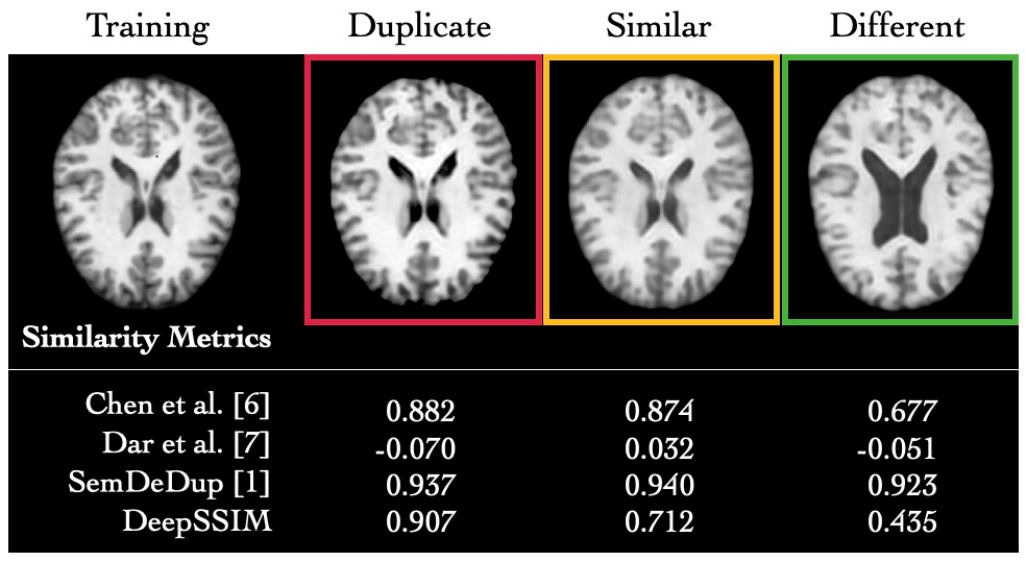}
\caption{
Example from the Brain MRI dataset. The figure shows a real training image alongside three synthetic counterparts generated by the LDM and manually labeled by our experts.
}
\label{fig:benchmark-brain}
\end{figure}

\begin{figure}[t]
\centering
\includegraphics[width=\linewidth]{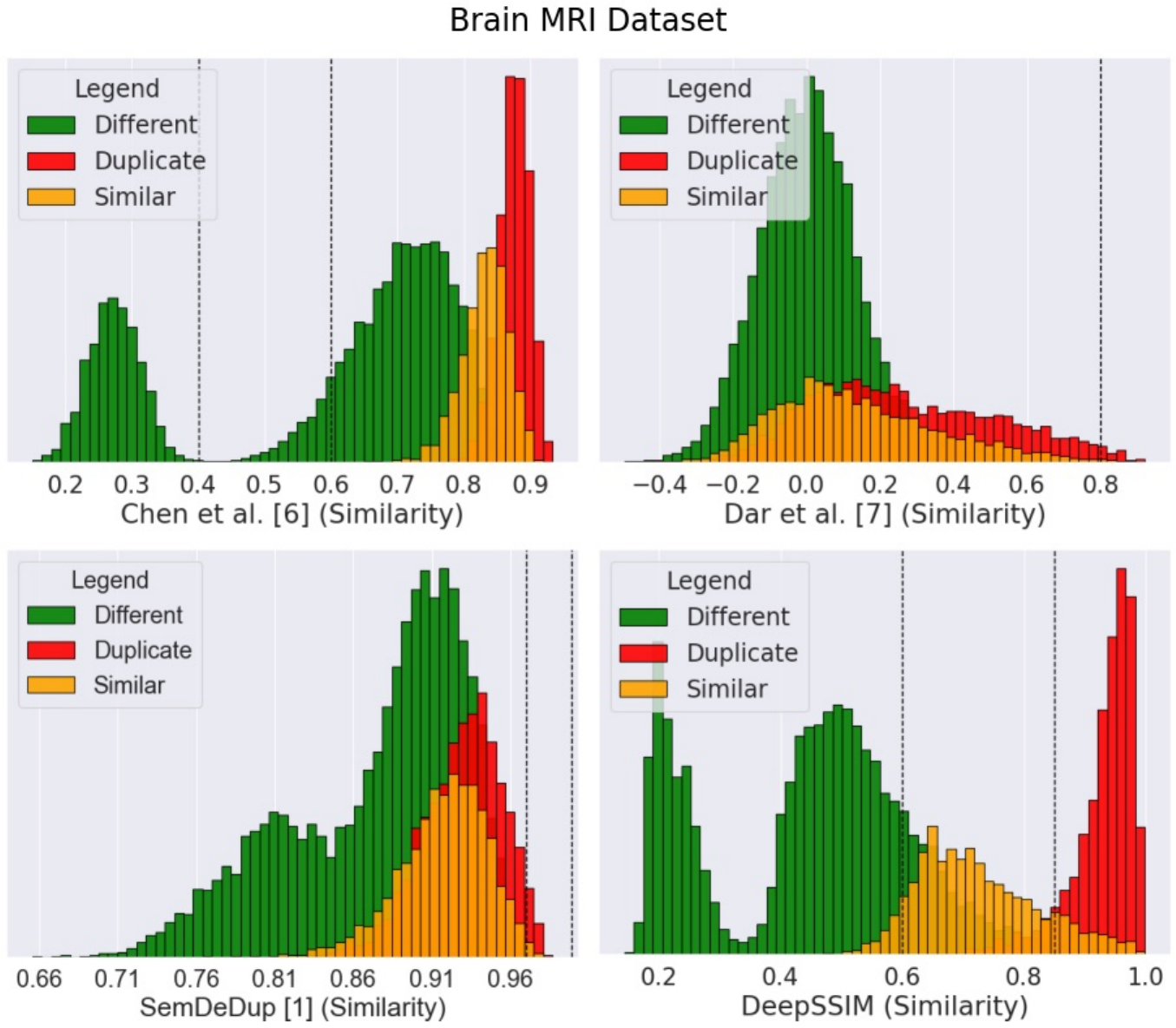}
\caption{{The histograms show the distribution of similarity scores for image pairs in the Brain MRI dataset, separated by class, across all competing methods. The vertical lines represent the classification thresholds applied by each method.}}
\label{fig:qualitative_compare_brain}
\end{figure}

\subsubsection{Performance on Aligned Pairs}
As shown in Table~\ref{tab:brainmri} (A), the superiority of DeepSSIM is most clearly highlighted by the Silhouette Score, which assesses class separability in each method independent of any threshold selection. DeepSSIM achieves the highest Silhouette Score (0.35) among all approaches, second only to the ideal SSIM (0.54). In contrast, other baselines yield Silhouette Scores near zero or negative, clearly indicating a lack of discriminative power for these solutions. The improved separation achieved by DeepSSIM results also in a high F1 score (81.55\%), surpassing all competing methods except for SSIM (99.98\%). However, the near-perfect performance of the SSIM baseline in this case is likely due to the images being perfectly aligned. 
 
\subsubsection{Performance on Spatially Misaligned Pairs}
As shown in Table~\ref{tab:brainmri} (B), DeepSSIM achieves the highest Silhouette Score (0.34) among all methods, including SSIM (0.04), demonstrating robust class separability despite spatial variations in the image pairs. Correspondingly, DeepSSIM maintains stable classification performance compared to all other baselines, attaining the highest macro F1 score of 81.00\%. Notably, the SSIM baseline suffers a severe decline, with its F1 score dropping to 30.32\%, highlighting a critical vulnerability to minor spatial misalignments common in real-world scenarios.

\begin{table*}[t]
\centering
\setlength{\tabcolsep}{5pt}
\def\arraystretch{1.25}

\resizebox{0.95\textwidth}{!} {
\begin{tabular}{|c|cc|cc|cc|c|c|}
\hline
\multirow{2}{*}{\textbf{Method}} 
& \multicolumn{2}{c|}{\textbf{Different}} 
& \multicolumn{2}{c|}{\textbf{Similar}} 
& \multicolumn{2}{c|}{\textbf{Duplicate}} 
& \multirow{2}{*}{\textbf{Macro F1 score}} 
& \multirow{2}{*}{\textbf{Silhouette Score}} \\
\cline{2-7}
& \textbf{Precision} & \textbf{Recall} 
& \textbf{Precision} & \textbf{Recall} 
& \textbf{Precision} & \textbf{Recall} 
& & \\
\hline

\multicolumn{9}{|c|}{\textbf{(A) Performance on Aligned Pairs}} \\
\hline
Chen et al.~\cite{chen2024extracting} & \best{100.00\%} & 5.04\% & 0.00\% & 0.00\% & 11.42\% & \best{100.00\%} & 10.03\% & 0.28 \\
Dar et al.~\cite{dar2023investigating} & 71.68\% & 99.84\% & N/A & N/A & 42.03\% & 11.20\% & 33.71\% & 0.25 \\
SemDeDup~\cite{abbas2023semdedup} & 74.79\% & \second{99.92\%} & 52.62\% & 14.10\% & 0.00\% & 0.00\% & 35.93\% & 0.09 \\
SSIM & \best{100.00\%} & \best{100.00\%} & \best{99.88\%} & \best{100.00\%} & \best{100.00\%} & \second{99.66\%} & \best{99.92\%} & \best{0.53} \\
DeepSSIM & \second{91.70\%} & 98.93\% & \second{85.57\%} & \second{54.66\%} & \second{61.29\%} & 82.00\% & \second{77.34\%} & \second{0.44} \\
\hline

\multicolumn{9}{|c|}{\textbf{(B) Performance on Spatially Misaligned Pairs}} \\
\hline
Chen et al.~\cite{chen2024extracting} & \best{100.00\%} & 8.36\% & 1.30\% & 2.35\% & 13.53\% & \best{98.98\%} & 13.64\% & \second{0.20} \\
Dar et al.~\cite{dar2023investigating} & 70.91\% & \second{99.94\%} & N/A & N/A & 35.93\% & 3.90\% & 30.00\% & 0.17 \\
SemDeDup~\cite{abbas2023semdedup} & 70.86\% & \best{100.00\%} & 62.96\% & 1.91\% & 0.00\% & 0.00\% & 28.88\% & 0.02 \\
SSIM & 71.46\% & \best{100.00\%} & \second{66.07\%} & \second{4.15\%} & \best{100.00\%} & 1.52\% & \second{31.39\%} & 0.16 \\
DeepSSIM & \second{91.81\%} & 98.18\% & \best{80.00\%} & \best{57.97\%} & \second{62.35\%} & \second{72.83\%} & \best{76.43\%} & \best{0.42} \\
\hline
\end{tabular}
}
\caption{ Quantitative performance evaluation using precision, recall, macro F1 score, and Silhouette Score on \textbf{Chest X-ray} data. Results are reported for: \textbf{(A) Aligned Pairs}, consisting of perfectly aligned image pairs; and \textbf{(B) Spatially Misaligned Pairs}, consisting of augmented images with simulated real-world spatial variability.}
\label{tab:chestxray}
\end{table*}

\subsection{Qualitative Comparison}
Figure~\ref{fig:qualitative_compare_brain} presents the distribution of similarity scores obtained by each method across the different classes. While the baseline methods struggle to distinguish between classes, our proposed approach produces well-separated score distributions, in line with the Silhouette Score results reported in Table~\ref{tab:brainmri}. Furthermore, Figure~\ref{fig:benchmark-brain} provides a visual example, showing one real brain MRI alongside synthetic images labeled as \textit{duplicate}, \textit{similar}, and \textit{different}. The figure also reports the similarity scores assigned by each method when comparing these images to the real scan. Consistent with the quantitative results, this example further illustrates the superior discriminative capability of our approach over competing methods.

\begin{figure}[t]
\centering
\includegraphics[width=\linewidth]{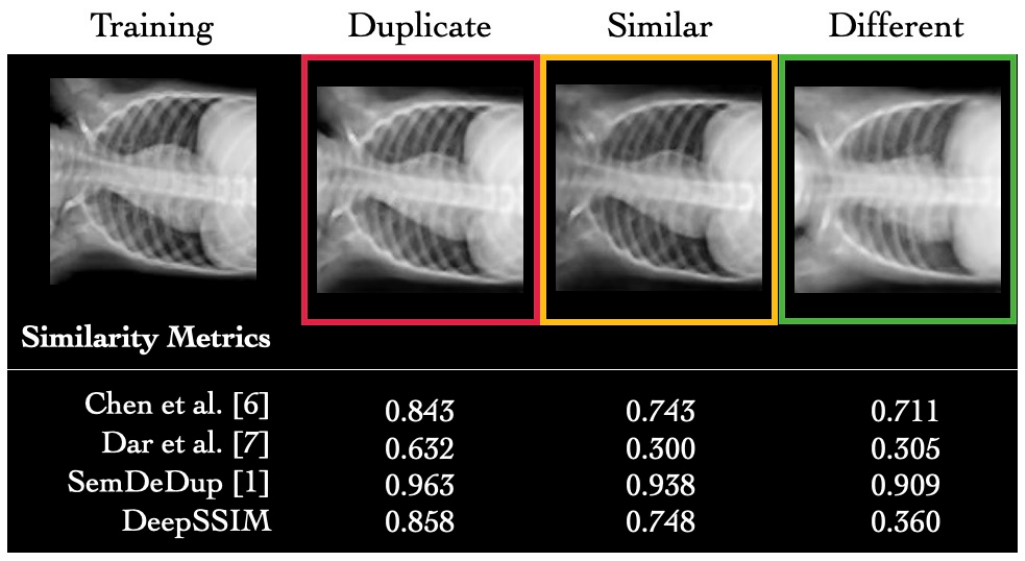}
\caption{
Example from the Chest X-ray dataset. The figure shows a real training image alongside three synthetic counterparts generated by the LDM and manually labeled by our experts.
}
\label{fig:benchmark-chest}
\end{figure}

\begin{figure}[t]
\centering
\includegraphics[width=\linewidth]{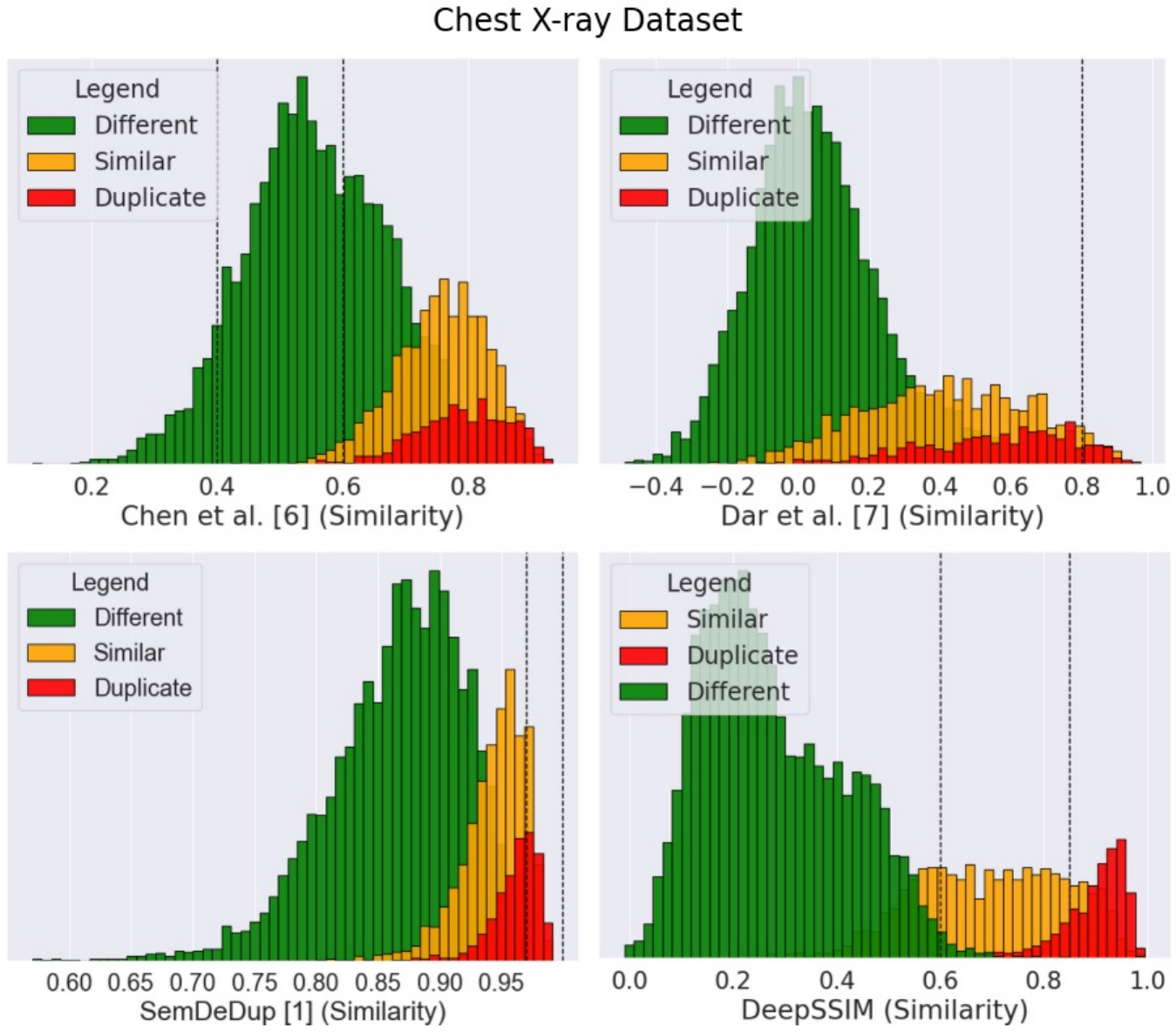}
\caption{ The histograms show the distribution of similarity scores for image pairs in the Chest X-ray dataset, separated by class, across all competing methods. The vertical lines represent the classification thresholds applied by each method.}
\label{fig:qualitative_compare_chest}
\end{figure}

\subsection{Generalization on the Chest X-ray Dataset}
To assess the generalizability of our approach, we test DeepSSIM on a different imaging modality and anatomical region. Specifically, we use 2,639 scans from the Pneumonia Chest X-ray dataset~\cite{kermany2018labeled}. For this experiment, we train a separate LDM on the new dataset and follow the procedure outlined in Section~\ref{sec:synthset}, resulting in 8,000 labeled image pairs for evaluation. We then re-train both DeepSSIM and Dar et al.~\cite{dar2023investigating} on this dataset using the same protocol as in the brain MRI experiments. The other methods rely on pre-trained models and therefore do not require retraining.

\subsubsection{Evaluation Results on Chest X-ray}
Across both the aligned-pairs setting (Table~\ref{tab:chestxray} (A)) and the more challenging setting with spatial variations (Table~\ref{tab:chestxray} (B)), DeepSSIM exhibits performance trends consistent with those observed on brain MRI data. Specifically, it obtains the second-highest macro F1 score (77.34\%) and Silhouette score (0.44) in the aligned image pairs setting, and outperforms all methods by achieving the highest F1 score (76.43\%) and Silhouette Score (0.42) in the spatially misaligned setting. These quantitative results are supported by Figure~\ref{fig:qualitative_compare_chest}, where DeepSSIM shows better-separated similarity score distributions, in contrast to the more overlapping distributions of the baseline methods. Finally, Figure~\ref{fig:benchmark-chest} presents a qualitative example of a real chest X-ray alongside synthetic images from the three classes, visually highlighting DeepSSIM’s effectiveness on this new modality. We believe our method remains effective on this dataset because its imaging context closely resembles that of brain MRI scans, consistently capturing a fixed anatomical region across patients. Similar to brain MRI, duplicates in this dataset are characterized by anatomy-preserving transformations, whereas non-duplicate image pairs exhibit visible structural variations.

\subsection{Thresholds Sensitivity Analysis}
\begin{figure}[h!]
\centering
\includegraphics[width=0.46\textwidth]{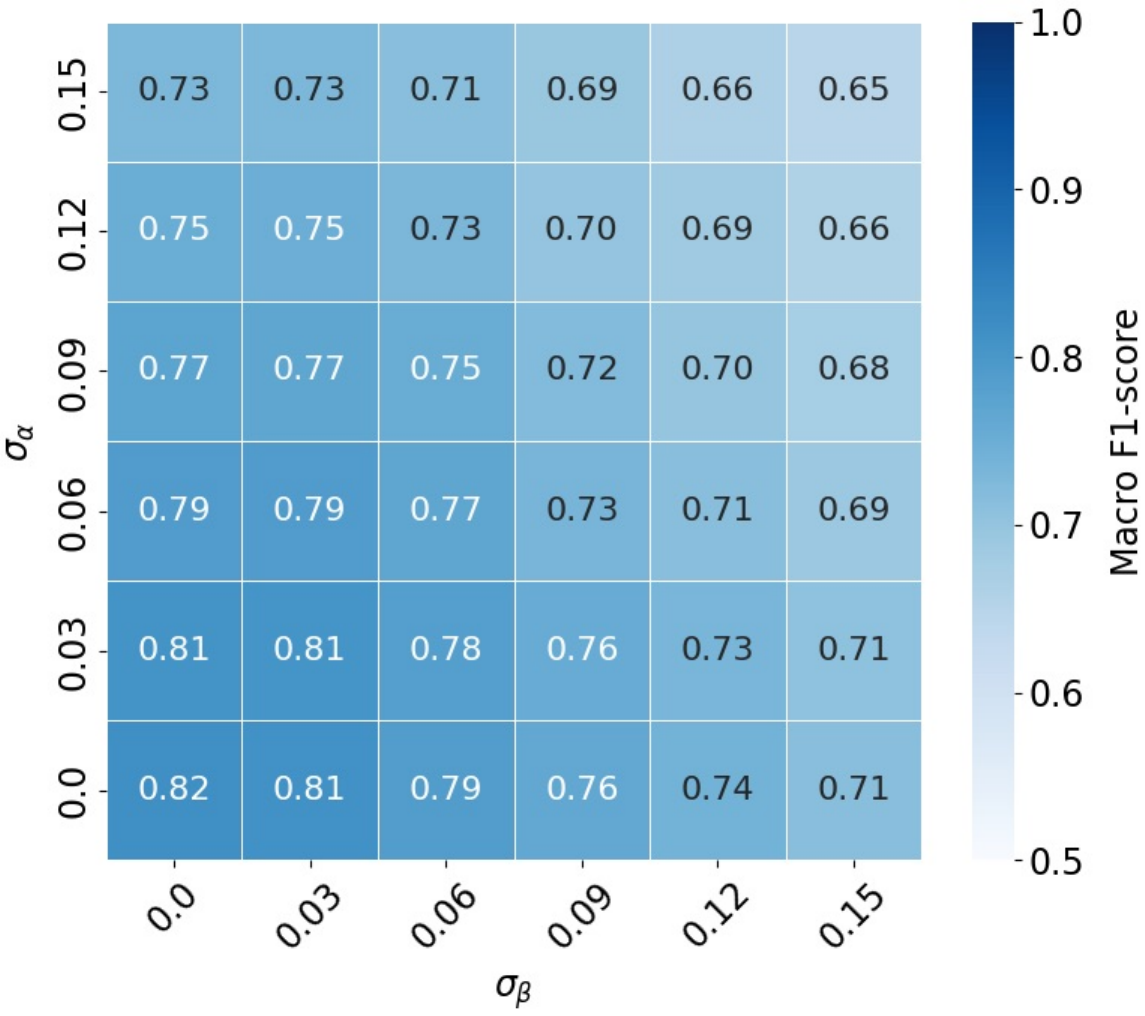}
\caption{ Heatmap of the F1 score as a function of threshold perturbations. The $y$- and $x$-axes correspond to increasing noise levels applied to $\alpha$ and $\beta$, respectively.}
\label{fig:thresholds}
\end{figure}

To evaluate the robustness of thresholds $\alpha$ and $\beta$ in our DeepSSIM pipeline, we examine their sensitivity to stochastic perturbations using the brain MRI test set. Starting from the selected values $\alpha=0.6$ and $\beta=0.85$, we add zero-mean Gaussian noise to each threshold, with standard deviations $\sigma_\alpha$ and $\sigma_\beta$ respectively. We measure how the F1 score changes when the noise has increasing standard deviations $\sigma_\alpha,\sigma_\beta$ from 0.03 to 0.15 (in increments of 0.03). As shown in Figure~\ref{fig:thresholds}, the F1 score decreases gradually with increasing noise, but the decline remains limited. Importantly, even at the highest noise levels, the performance consistently exceeds all baselines reported in Table~\ref{tab:brainmri}.

\subsection{Runtime Efficiency}\label{sec:runtime}
Beyond its superior performance, DeepSSIM holds a significant computational advantage over traditional SSIM. By encoding images into compact vector embeddings, our method can leverage highly optimized nearest neighbor search frameworks (e.g., FAISS~\cite{johnson2019billion}). To empirically demonstrate these advantages, we compare the runtime performance of SSIM and DeepSSIM, both executed on the same \textsc{NVIDIA RTX 4070 Ti Super} GPU. DeepSSIM completes embedding all images from datasets $\mathcal{D}$ and $\mathcal{S}$ (see Section~\ref{sec:synthset}) in 1 minute and 5 seconds and performs all vector similarity searches using FAISS in just 2 seconds. In contrast, SSIM requires approximately 596 minutes to complete the equivalent tasks, highlighting DeepSSIM’s superior efficiency and scalability.

\section{Conclusion}
\label{sec:conclusion}
In this study, we introduced DeepSSIM, a novel metric for evaluating memorization in deep generative models. By effectively capturing fine-grained anatomical details, DeepSSIM is particularly well-suited for analyzing memorization in medical imaging contexts. Our comparison with state-of-the-art methods demonstrates that DeepSSIM achieves superior accuracy in correctly identifying duplicates. We envision multiple applications for this method, as it can serve as a robust metric to reassess memorization across various configurations of generative models. This could lead to a more precise understanding of model behavior and its implications in medical and other sensitive domains. Additionally, some privacy-preserving techniques may fail to prevent memorization risks, as models can still retain sensitive information. Therefore, memorization detection techniques like DeepSSIM are essential to safeguard patient information. In conclusion, we believe our solution provides a scalable and robust method to detect and quantify memorization, ensuring that model training does not compromise patient confidentiality, data security, and privacy.

\section*{Acknowledgments}
LP is supported by the PNRR initiative (DM 118/2023).\\
The work of FG and SB has been supported by MUR in the framework of PNRR PE0000013, under project “Future Artificial Intelligence Research – FAIR”.\\
DR acknowledges funding from the “Rete eHealth: AI e strumenti ICT Innovativi orientati alla Diagnostica Digitale (RAIDD)” project (CUP J43C22000380001)).

{
\small
\bibliographystyle{ieeenat_fullname}
\bibliography{main}

@String(CVPR= {IEEE Conf. Comput. Vis. Pattern Recog.})

@String(ICLR = {Int. Conf. Learn. Represent.})

@String(CVPR  = {CVPR})

@String(ICLR  = {ICLR})

@article{webster2023reproducible,
  title={A reproducible extraction of training images from diffusion models},
  author={Webster, Ryan},
  journal={arXiv preprint arXiv:2305.08694},
  year={2023}
}

@article{chen2024extracting,
  title={Extracting Training Data from Unconditional Diffusion Models},
  author={Chen, Yunhao and Ma, Xingjun and Zou, Difan and Jiang, Yu-Gang},
  journal={arXiv preprint arXiv:2406.12752},
  year={2024}
}

@inproceedings{carlini2023extracting,
  title={Extracting training data from diffusion models},
  author={Carlini, Nicolas and Hayes, Jamie and Nasr, Milad and Jagielski, Matthew and Sehwag, Vikash and Tramer, Florian and Balle, Borja and Ippolito, Daphne and Wallace, Eric},
  booktitle={32nd USENIX Security Symposium (USENIX Security 23)},
  pages={5253--5270},
  year={2023}
}

@article{zhang2023biomedclip,
  title={Biomedclip: a multimodal biomedical foundation model pretrained from fifteen million scientific image-text pairs},
  author={Zhang, Sheng and Xu, Yanbo and Usuyama, Naoto and Xu, Hanwen and Bagga, Jaspreet and Tinn, Robert and Preston, Sam and Rao, Rajesh and Wei, Mu and Valluri, Naveen and others},
  journal={arXiv preprint arXiv:2303.00915},
  year={2023}
}

@inproceedings{somepalli2023diffusion,
  title={Diffusion art or digital forgery? investigating data replication in diffusion models},
  author={Somepalli, Gowthami and Singla, Vasu and Goldblum, Micah and Geiping, Jonas and Goldstein, Tom},
  booktitle={Proceedings of the IEEE/CVF Conference on Computer Vision and Pattern Recognition},
  pages={6048--6058},
  year={2023}
}

@inproceedings{dar2023investigating,
  title={Investigating data memorization in 3d latent diffusion models for medical image synthesis},
  author={Dar, Salman Ul Hassan and Ghanaat, Arman and Kahmann, Jannik and Ayx, Isabelle and Papavassiliu, Theano and Schoenberg, Stefan O and Engelhardt, Sandy},
  booktitle={International Conference on Medical Image Computing and Computer-Assisted Intervention},
  pages={56--65},
  year={2023},
  organization={Springer}
}

@inproceedings{chen2024towards,
  title={Towards Memorization-Free Diffusion Models},
  author={Chen, Chen and Liu, Daochang and Xu, Chang},
  booktitle={Proceedings of the IEEE/CVF Conference on Computer Vision and Pattern Recognition},
  pages={8425--8434},
  year={2024}
}

@inproceedings{pizzi2022self,
  title={A self-supervised descriptor for image copy detection. 2022 IEEE},
  author={Pizzi, Ed and Roy, Sreya Dutta and Ravindra, Sugosh Nagavara and Goyal, Priya and Douze, Matthijs},
  booktitle={CVF Conference on Computer Vision and Pattern Recognition (CVPR)},
  pages={14512--14522},
  year={2022}
}

@inproceedings{puglisi2024deepbrainprint,
  title={DeepBrainPrint: A Novel Contrastive Framework for Brain MRI Re-Identification},
  author={Puglisi, Lemuel and Eshaghi, Arman and Parker, Geoff and Barkhof, Frederik and Alexander, Daniel C and Ravi, Daniele},
  booktitle={Medical Imaging with Deep Learning},
  pages={716--729},
  year={2024},
  organization={PMLR}
}

@inproceedings{rombach2022high,
  title={High-resolution image synthesis with latent diffusion models},
  author={Rombach, Robin and Blattmann, Andreas and Lorenz, Dominik and Esser, Patrick and Ommer, Bj{\"o}rn},
  booktitle={Proceedings of the IEEE/CVF conference on computer vision and pattern recognition},
  pages={10684--10695},
  year={2022}
}

@article{gu2021domain,
  title={Domain-specific language model pretraining for biomedical natural language processing},
  author={Gu, Yu and Tinn, Robert and Cheng, Hao and Lucas, Michael and Usuyama, Naoto and Liu, Xiaodong and Naumann, Tristan and Gao, Jianfeng and Poon, Hoifung},
  journal={ACM Transactions on Computing for Healthcare (HEALTH)},
  volume={3},
  number={1},
  pages={1--23},
  year={2021},
  publisher={ACM New York, NY}
}

@article{somepalli2023understanding,
  title={Understanding and mitigating copying in diffusion models},
  author={Somepalli, Gowthami and Singla, Vasu and Goldblum, Micah and Geiping, Jonas and Goldstein, Tom},
  journal={Advances in Neural Information Processing Systems},
  volume={36},
  pages={47783--47803},
  year={2023}
}

@article{wang2004image,
  title={Image quality assessment: from error visibility to structural similarity},
  author={Wang, Zhou and Bovik, Alan C and Sheikh, Hamid R and Simoncelli, Eero P},
  journal={IEEE transactions on image processing},
  volume={13},
  number={4},
  pages={600--612},
  year={2004},
  publisher={IEEE}
}

@inproceedings{pinaya2022brain,
  title={Brain imaging generation with latent diffusion models},
  author={Pinaya, Walter HL and Tudosiu, Petru-Daniel and Dafflon, Jessica and Da Costa, Pedro F and Fernandez, Virginia and Nachev, Parashkev and Ourselin, Sebastien and Cardoso, M Jorge},
  booktitle={MICCAI Workshop on Deep Generative Models},
  pages={117--126},
  year={2022},
  organization={Springer}
}

@article{ho2020denoising,
  title={Denoising diffusion probabilistic models},
  author={Ho, Jonathan and Jain, Ajay and Abbeel, Pieter},
  journal={Advances in neural information processing systems},
  volume={33},
  pages={6840--6851},
  year={2020}
}

@article{johnson2019billion,
  title={Billion-scale similarity search with {GPUs}},
  author={Johnson, Jeff and Douze, Matthijs and J{\'e}gou, Herv{\'e}},
  journal={IEEE Transactions on Big Data},
  volume={7},
  number={3},
  pages={535--547},
  year={2019},
  publisher={IEEE}
}

@article{goodfellow2020generative,
  title={Generative adversarial networks},
  author={Goodfellow, Ian and Pouget-Abadie, Jean and Mirza, Mehdi and Xu, Bing and Warde-Farley, David and Ozair, Sherjil and Courville, Aaron and Bengio, Yoshua},
  journal={Communications of the ACM},
  volume={63},
  number={11},
  pages={139--144},
  year={2020},
  publisher={ACM New York, NY, USA}
}

@inproceedings{nagarajan2018theoretical,
  title={Theoretical insights into memorization in gans},
  author={Nagarajan, Vaishnavh and Raffel, Colin and Goodfellow, Ian J},
  booktitle={Neural Information Processing Systems Workshop},
  volume={1},
  pages={3},
  year={2018}
}

@inproceedings{ Liu_2022_CVPR,
    author    = {Liu, Zhuang and Mao, Hanzi and Wu, Chao-Yuan and Feichtenhofer, Christoph and Darrell, Trevor and Xie, Saining},
    title     = {A ConvNet for the 2020s},
    booktitle = {Proceedings of the IEEE/CVF Conference on Computer Vision and Pattern Recognition (CVPR)},
    month     = {June},
    year      = {2022},
    pages     = {11976-11986}
}

@inproceedings{ iqa_measures,
    title="A Study on the Adequacy of Common IQA Measures for Medical Images",
    author="Breger, Anna
    and Karner, Clemens
    and Selby, Ian
    and Gr{\"o}hl, Janek
    and Dittmer, S{\"o}ren
    and Lilley, Edward
    and Babar, Judith
    and Beckford, Jake
    and Else, Thomas R.
    and Sadler, Timothy J.
    and Shahipasand, Shahab
    and Thavakumar, Arthikkaa
    and Roberts, Michael
    and Sch{\"o}nlieb, Carola-Bibiane",
    year="2025",
    publisher="Springer Nature Singapore",
    pages="451--462",
    isbn="978-981-96-3863-5"
}

@inproceedings{abbas2023semdedup,
  title={SemDeDup: Data-efficient learning at web-scale through semantic deduplication},
  year={2023},
  author={Abbas, Amro Kamal Mohamed and Tirumala, Kushal and Simig, Daniel and Ganguli, Surya and Morcos, Ari S},
  booktitle={ICLR 2023 Workshop on Multimodal Representation Learning: Perks and Pitfalls}
}

@inproceedings{radford2021learning,
  title={Learning transferable visual models from natural language supervision},
  author={Radford, Alec and Kim, Jong Wook and Hallacy, Chris and Ramesh, Aditya and Goh, Gabriel and Agarwal, Sandhini and Sastry, Girish and Askell, Amanda and Mishkin, Pamela and Clark, Jack and others},
  booktitle={International conference on machine learning},
  pages={8748--8763},
  year={2021},
  organization={PmLR}
}

@inproceedings{wen2024detecting,
  title={Detecting, explaining, and mitigating memorization in diffusion models},
  author={Wen, Yuxin and Liu, Yuchen and Chen, Chen and Lyu, Lingjuan},
  booktitle={The Twelfth International Conference on Learning Representations},
  year={2024}
}

@inproceedings{jiang2025image,
  title={Image-level Memorization Detection via Inversion-based Inference Perturbation},
  author={Jiang, Yue and Lin, Haokun and Bai, Yang and Peng, Bo and Liu, Zhili and Lyu, Yueming and Yang, Yong and Dong, Jing and others},
  booktitle={The Thirteenth International Conference on Learning Representations},
  year={2025}
}

@inproceedings{shi2024detecting,
title={Detecting Pretraining Data from Large Language Models},
author={Weijia Shi and Anirudh Ajith and Mengzhou Xia and Yangsibo Huang and Daogao Liu and Terra Blevins and Danqi Chen and Luke Zettlemoyer},
booktitle={The Twelfth International Conference on Learning Representations},
year={2024},
url={https://openreview.net/forum?id=zWqr3MQuNs}
}

@ARTICLE{fsim_5705575,
  author={Zhang, Lin and Zhang, Lei and Mou, Xuanqin and Zhang, David},
  journal={IEEE Transactions on Image Processing}, 
  title={FSIM: A Feature Similarity Index for Image Quality Assessment}, 
  year={2011},
  volume={20},
  number={8},
  pages={2378-2386},
  doi={10.1109/TIP.2011.2109730}}

@article{kermany2018labeled,
  title={Labeled optical coherence tomography (oct) and chest x-ray images for classification},
  author={Kermany, Daniel},
  journal={Mendeley data},
  year={2018},
  publisher={mendeley}
}

@inproceedings{chen2020gan,
  title={Gan-leaks: A taxonomy of membership inference attacks against generative models},
  author={Chen, Dingfan and Yu, Ning and Zhang, Yang and Fritz, Mario},
  booktitle={Proceedings of the 2020 ACM SIGSAC conference on computer and communications security},
  pages={343--362},
  year={2020}
}

@article{rousseeuw1987silhouettes,
  title={Silhouettes: a graphical aid to the interpretation and validation of cluster analysis},
  author={Rousseeuw, Peter J},
  journal={Journal of computational and applied mathematics},
  volume={20},
  pages={53--65},
  year={1987},
  publisher={Elsevier}
}
}

\end{document}